%% file: emnlp2021.tex
\pdfoutput=1
\documentclass[11pt]{article}

\usepackage{emnlp2021}

\usepackage{times}
\usepackage{latexsym}

\usepackage[T1]{fontenc}
\usepackage[utf8]{inputenc}

\usepackage{microtype}

\usepackage{smile}
\usepackage{subcaption}
\usepackage[ruled,vlined]{algorithm2e}
\newcommand{\ours}{\text{ARCH~}}

\usepackage{setspace}
\title{ARCH: Efficient Adversarial Regularized Training with Caching}

\author{
Simiao Zuo$^\ddagger$\thanks{\hspace{0.03in} Corresponding author.}, \ Chen Liang$^\ddagger$, \ Haoming Jiang$^\Box$\thanks{\hspace{0.03in} Work was done at Georgia Institute of Technology.}, \
Pengcheng He$^\diamond$, \ Xiaodong Liu$^\diamond$, \\
\textbf{Jianfeng Gao$^\diamond$, \ Weizhu Chen$^\diamond$ and Tuo Zhao$^{\ddagger}$} \\
$^\ddagger$Georgia Institute of Technology \ \
$^\Box$Amazon \ \
$^\diamond$Microsoft \\
\texttt{\{simiaozuo,cliang73\}@gatech.edu}, \ \texttt{jhaoming@amazon.com} \\
\texttt{\{Pengcheng.H,xiaodl,jfgao,wzchen\}@microsoft.com}, \\ \texttt{tourzhao@gatech.edu}
}

\begin{document}
\maketitle

\begin{abstract}
Adversarial regularization can improve model generalization in many natural language processing tasks.
However, conventional approaches are computationally expensive since they need to generate a perturbation for each sample in each epoch. 
We propose a new adversarial regularization method ARCH (adversarial regularization with caching), where perturbations are generated and cached once every several epochs.
As caching all the perturbations imposes memory usage concerns, we adopt a K-nearest neighbors-based strategy to tackle this issue. The strategy only requires caching a small amount of perturbations, without introducing additional training time.
We evaluate our proposed method on a set of neural machine translation and natural language understanding tasks. We observe that ARCH significantly eases the computational burden (saves up to 70\% of computational time in comparison with conventional approaches). More surprisingly, by reducing the variance of stochastic gradients, ARCH produces a notably better (in most of the tasks) or comparable model generalization. 
Our code is available at \url{https://github.com/SimiaoZuo/Caching-Adv}.
\end{abstract}

\input{0-introduction}
\input{0-background}
\input{0-method}

\input{0-experiments}
\input{0-conclusion}
% \input{0-broader-impact}

% \clearpage

\bibliography{anthology,custom}
\bibliographystyle{acl_natbib}

\clearpage
\appendix
\input{0-appendix}

\end{document}

%% file: 0-introduction.tex
\section{Introduction}

Adversarial regularization \cite{miyato2016adversarial} can improve model generalization in many natural language processing tasks, such as neural machine translation \cite{cheng2019robust}, natural language understanding \cite{jiang2019smart}, language modeling \cite{wang2019improving}, and reading comprehension \cite{jia2017adversarial}.
Even though the method has demonstrated its power in many scenarios, its computational efficiency remains unsatisfactory.

Conventional adversarial regularization~\cite{miyato2016adversarial} methods involve a min-max optimization problem. Specifically, a perturbation is generated for each sample by solving a maximization problem, and the model parameters are subsequently updated through a minimization problem, subject to the generated perturbations.
A popular algorithm~\cite{madry2017towards} for such optimization is to alternate between several projected gradient descent steps (PGD, for the maximization) and a gradient descent step (for the minimization).

There are two drawbacks with the alternating gradient descent/ascent method.
First, the procedure requires significant computational efforts. Suppose we run PGD for $S$ steps, then we introduce extra $S$ forward passes and extra $S$ backward passes in each iteration. As such, training with adversarial regularization is significantly slower than standard training.
Second, optimizing the min-max problem is hard. This is because the perturbations are model and data dependent, and thus, variance of them is large. That is, the model needs to adapt to drastically different ``noisy data'' (i.e., clean data with perturbations), such that the stochastic gradients vary significantly during training. Such large variance imposes optimization challenges.
% \citet{aghajanyan2020better} replace the perturbations with random noise to resolve the efficiency issue. However, the method exhibits large run-to-run variance, and the optimization problem still exists.

We propose \ours (\textbf{A}dversarial \textbf{R}egularization with \textbf{C}ac\textbf{H}ing) that alleviates the aforementioned issues by reusing perturbations. Recall that in conventional adversarial regularization methods, a different perturbation is generated for each sample in each epoch. In contrast to this, we propose to generate perturbations less frequently. For example, for a given sample, we can generate a new perturbation every 20 epochs, and the sample's perturbation remains unchanged in other epochs.
We call this method ``\textit{caching}''.
The method has two advantages. First, it alleviates the computational burden. By reusing the perturbations, we avoid the extra forward and backward passes caused by PGD for most of the iterations.
Second, caching stabilizes the stochastic gradients. Notice that in our method, the model is optimized with respect to the same noisy data for multiple times, instead of only one.
In this way, variance of the stochastic gradients is reduced.

One caveat of the caching method is its memory overhead. This is because a sample's perturbation is significantly larger than itself (the perturbation has an extra embedding dimension). We propose a K-nearest neighbors-based approach to tackle this problem.
Specifically, instead of caching perturbations for all the samples, we only cache a small proportion of them. Each uncached perturbation can then be constructed using the cached ones in its neighborhood. Such a construction procedure can be executed in parallel with model training. Therefore, training time will not be prolonged because of this memory saving strategy.

We use a moving average approach to boost model generalization. Specifically, when generating a new perturbation, we integrate information from both the current model and the current perturbation. This is different from conventional approaches, where the new perturbation only depends on the current model.
The moving average approach has a smoothing effect that boosts model generalization, as demonstrated both theoretically and empirically by previous works~\cite{izmailov2018averaging, athiwaratkun2018there, jiang2019smart}.

Arguably, the perturbations introduced by our method may not constitute strong adversarial attacks, because of the ``staleness'' caused by infrequent updates. However, we highlight that the focus of this work is \textbf{model generalization} over clean data, instead of adversarial robustness (ability to defend attacks). As we will demonstrate in the experiments, the ``weak'' perturbations show notable improvement of model generalization. And somewhat surprisingly, \ours also exhibits on par or even better robustness comparing with conventional approaches.

We conduct extensive experiments on neural machine translation (NMT) and natural language understanding (NLU) tasks. In comparison with conventional adversarial regularization approaches, \ours can save up to 70\% computational time. Moreover, in NMT tasks, our method improves about 0.5 BLUE over baseline methods on seven datasets. \ours also achieves 0.7 average score improvement on the GLUE~\cite{wang2018glue} development set over existing methods.

We summarize our contributions as follows:
(1) We propose a caching method that needs drastically less computational efforts. The method can also improve model generalization by reducing variance of stochastic gradient.
(2) We propose a memory saving strategy to efficiently implement the caching method.
(3) Extensive experiments on neural machine translation and natural language understanding demonstrate the efficiency and effectiveness of the proposed method.

%% file: 0-background.tex
\vspace{-0.05in}
\section{Background}
\vspace{-0.05in}

%%%%%%%%%%%%%%%%%%%%%%%%%%%%%%
\noindent $\diamond$
\textbf{Neural machine translation} has achieved superior empirical performance \cite{bahdanau2014neural, gehring2017convolutional, vaswani2017attention}. Recently, the Transformer \cite{vaswani2017attention} architecture dominates the field. This sequence-to-sequence model employs an encoder-decoder structure, and also integrates the attention mechanism.
During the encoding phase, a Transformer model first computes an embedding for each sentence, after which the embeddings are fed into several layers of encoding blocks. Each of these blocks contain a self-attention mechanism and a feed-forward neural network (FFN). Subsequently, after encoding, the hidden representations are fed into the decoding blocks, each constituted of a self-attention, a encoder-decoder attention, and a FFN.

%%%%%%%%%%%%%%%%%%%%%%%%%%%%%%
% \vspace{0.05in}
\noindent $\diamond$
\textbf{Fine-tuning pre-trained language models} \cite{peters2018deep, devlin2018bert, radford2019language, liu2019roberta, he2020deberta} is a state-of-the-art method for natural language understanding tasks such as the GLUE \cite{wang2018glue} benchmark.
Adversarial regularization is also incorporated into the fine-tuning approach. For example, \citet{liu2020adversarial} combines adversarial pre-training and fine-tuning, \citet{zhu2019freelb, jiang2019smart} adopt trust region-based methods, and \citet{aghajanyan2020better} aims for a more efficient computation.

%%%%%%%%%%%%%%%%%%%%%%%%%%%%%%
% \vspace{0.05in}
\noindent $\diamond$
\textbf{Adversarial training} was originally proposed for computer vision tasks~\cite{szegedy2013intriguing, goodfellow2014explaining, madry2017towards}, where the goal is to train robust classifiers. Such methods synthesize adversarial samples, such that the classifier is trained to be robust against them.
This strategy is also effective for tasks beyond computer vision, such as in reinforcement learning \cite{shen2020deep}.
Various algorithms are proposed to craft the adversarial samples, e.g., learning-to-learn \cite{jiang2021learning} and Stackelberg adversarial training \cite{zuo2021adversarial}. Moreover, adversarial training is also well-studied theoretically \cite{li2019inductive}.
In natural language processing, the goal is no longer adversarial robustness, but instead we use adversarial regularization to boost model generalization. Note that adversarial training and adversarial regularization are different concepts. The former focuses on defending against adversarial attacks, and the latter focuses on encouraging smooth model predictions~\cite{miyato2016adversarial}. These two goals are usually treated as mutually exclusive~\cite{raghunathan2020understanding, min2020curious}.

%% file: 0-method.tex
\section{Method}

%%%%%%%%%%%%%%%%%%%%%%%%%%%%%%
Generating perturbations for natural language inputs faces the difficulty of discreteness, i.e., words are defined in a discrete space.
A common approach to tackle this is to work on the continuous embedding space \cite{miyato2016adversarial, sano2019effective}.
Denote $f(x,\theta)$ a neural network parameterized by $\theta$, where $x$ is the input embedding.
Further denote $y$ the ground-truth corresponding to $x$.
For example, in classification tasks, $x$ is the sentence embedding, and $y$ is its label. In sequence-to-sequence learning, $x$ is the source sentence embedding, and $y$ is the target sentence.
In both of these cases, the model is trained by minimizing the empirical risk over the training data, i.e.,
\begin{equation*}
    \min_{\theta} \cL(\theta)=\frac{1}{n} \sum_{i=1}^n \ell( f(x_i,\theta), y_i ).
\end{equation*}
Here $\{(x_i,y_i)\}_{i=1}^n$ is the dataset, and $\ell$ is a task-specific loss, e.g., cross-entropy loss for classification and mean-squared error for regression.

%%%%%%%%%%%%%%%%%%%%%%%%%%%%%%
\subsection{Adversarial Regularization}

Adversarial regularization \cite{miyato2016adversarial} is a technique that encourages smoothness of the model outputs around each input data point.
Concretely, we define an adversarial regularizer for non-regression tasks as
\begin{align*}
    &\ell_{v}(x, \delta, \theta) = \mathrm{KL}\big( f(x, \theta) ~||~ f(x + \delta, \theta) \big), \\[5pt]
    &\text{where } \mathrm{KL}(P ~||~ Q) = \sum_k p_k \log \frac{p_k}{q_k}.
\end{align*}
Here $f(\cdot, \theta)$ is the prediction confidence, i.e., $\sum_i [f(\cdot,\theta)]_i=1$, $\delta$ is the perturbation of sample $x$, and $\mathrm{KL}(\cdot || \cdot)$ is the Kullback–Leibler (KL) divergence.
In regression tasks, the model output $f(\cdot,\theta)$ is a scalar, and the adversarial regularizer is
\begin{equation*}
    \ell_v \left(x, \delta, \theta) = (f(x,\theta) - f(x+\delta,\theta) \right)^2.
\end{equation*}

We consider the worst-case perturbation to encourage the model to make smooth predictions. Specifically, at epoch $t$, we solve
\begin{equation} \label{eq:vat-loss}
    \min_{\theta^t} \cL(\theta^t)
    + \frac{\lambda}{n} \sum_{i=1}^n\max_{\norm{\delta_i^t} \leq \epsilon} \ell_{v}( x_i, \delta_i^t, \theta^t ).
\end{equation}
Here $\lambda$ is the weight of the regularizer, $\epsilon$ is a pre-defined perturbation strength, and $\norm{\cdot}$ is either the $\ell_2$ norm or the $\ell_\infty$ norm. Notice that the perturbation $\delta_i^t$ of sample $x_i$ is different in each epoch.

The min-max optimization problem in Eq.~\ref{eq:vat-loss} is notoriously difficult to solve. Previous works~\cite{miyato2016adversarial, sano2019effective, jiang2019smart, zhu2019freelb} employ variations of alternating gradient descent/ascent. That is, we first solve the maximization problem using several iterations of projected gradient ascent, and then we run a gradient descent step on the loss function of the minimization problem, subject to the generated perturbations. The above procedures are run iteratively.

On major drawback of the alternating gradient descent/ascent approach is that the \textbf{stochastic gradients are unstable}. Specifically, norms of the gradients vary significantly during training (Fig.~\ref{fig:grad-norm}).
This is because perturbations are generated based on the current model parameters, i.e., by maximizing $\ell_v(x_i, \delta_i^t, \theta^t)$, where $\theta^t$ changes in each epoch. Therefore, the perturbations exhibit large variance. This causes instability of the stochastic gradients, because the model needs to adapt to drastically different adversarial directions (i.e., $\delta_i^t$).

%%%%%%%%%%%%%%%%%%%%%%%%%%%%%%%%%%%%%%%%
\subsection{Adversarial Regularization with Caching}

To alleviate the gradient instability problem, we propose to reuse the perturbations. Specifically, instead of optimizing with respect to different perturbations $\{\delta_i^t\}_{i=1}^n$ in each epoch, we optimize with respect to the same ones for several epochs.

Concretely, the training objective is now
\begin{align} \label{eq:cache-loss}
    &\min_{\theta^t} \cL(\theta^t)
    + \frac{\lambda}{n} \sum_{i=1}^n \ell_{v}( x_i, \delta_i^t, \theta^t ), \\
    &\delta_i^t =
    \begin{cases}
        \delta_i^{t-1}, & t \% T_c \neq 0, \\
        \alpha \delta_i^{t-1} + (1-\alpha) \Delta_i^t, & t \% T_c = 0.
    \end{cases} \notag \\
    &\quad\text{where } \Delta_i^t = \max_{\norm{\delta_i^t} \leq \epsilon} \ell_{v}( x_i, \delta_i^t, \theta^t ). \notag
\end{align}
Here, $\%$ is the mod operator, and $T_c$ is a pre-defined gap between re-computing the perturbations.
Notice that we use an exponential moving average (EMA) approach with parameter $\alpha$ when updating the perturbations. The EMA strategy integrates past information into the current epoch, and induces a smoothing effect that boosts model generalization. This strategy has demonstrated its effectiveness in many previous works~\cite{izmailov2018averaging, athiwaratkun2018there, jiang2019smart}.

In comparison with Eq.~\ref{eq:vat-loss}, the formulation in Eq.~\ref{eq:cache-loss} indicates that the perturbations are generated $\lfloor T/T_c \rfloor$ times instead of $T$ times when we train for $T$ epochs. As such, the model is optimized with respect to $\{\delta_i\}_{i=1}^n$ for $T_c$ times, instead of only one time. In this way, the model can better adapt to the perturbed data, and thus, variance of the gradient norms is reduced. Intuitively, this is because optimization is more stable when the model is trained on the same data for multiple epochs, in comparison with trained on different noisy data in each epoch.
The algorithm to implement the caching strategy is summarized in Algorithm~\ref{alg:cache}.

In conventional adversarial regularization (e.g., SMART), we find the perturbations by optimization algorithms such as projected gradient decent at every iteration.
Recently, R3F~\cite{aghajanyan2020better} propose to use random perturbations instead, i.e., they directly draw $\delta$ from a normal distribution, and generalization of R3F can match SMART in some cases. However, because the random noise (as opposed to optimized perturbations) is not data-dependent, generalization of R3F is subpar in some scenarios, e.g., machine translation (see our experiments). Our approach enjoys the advantages of both of these two methods. Specifically, \ours is efficient since it remove the maximization problem most of the time. Moreover, perturbations generated by our method are informative, unlike R3F. Empirically, our proposed method is just as efficient as R3F, and somewhat surprisingly, we find that generalization of \ours can not only match, but even surpass conventional approaches in most of the tasks (see our experiments).

% By using Algorithm~\ref{alg:cache}, the generated perturbations may not constitute strong adversarial attacks. That is, for $t \% T_c \neq 0$, the perturbations are not computed by maximizing $\ell_v(x, \delta^t, \theta^t)$, and thus they may deviate from the optimal adversarial directions.
% In practice, however, increased bias does not hinder model generalization.
% For example, R3F~\cite{aghajanyan2020better} directly draws $\delta$ from a normal distribution, and its generalization can match SMART in many cases.

% For example, while conventional adversarial regularization (e.g., SMART) finds optimal perturbations at every iteration, R3F~\cite{aghajanyan2020better} directly draws $\delta$ from a normal distribution. Empirically, R3F is efficient and its generalization of it can match SMART in some cases.
% Our method is a trade-off between these two approaches, and somewhat surprisingly, we find that generalization of \ours can not only match, but even surpass SMART and R3F while being significantly faster than conventional approaches.

\begin{algorithm}[htb!]
\SetAlgoLined
\caption{Adversarial Regularization with Caching.}
\label{alg:cache}
\KwIn{$T$: number of training epochs; $T_c$: number of epochs between caching; $\alpha$: moving average parameter.}
\textbf{Initialize:} Cache $\cC=\text{dict}\{\}$\;
\For{$t=0, \cdots T-1$}{
    \For{each batch $\cB$}{
        \tcp{Find perturbations}
        \uIf{$t \% T_c == 0$}{
            Find $\delta_i^t$ for each $x_i \in \cB$ using projected gradient ascent\;
            $\cC[x_i] \leftarrow \alpha \cC[x_i] + (1-\alpha) \delta_i^t$ for each $x_i \in \cB$\;
        }
        \Else{
            $\delta_i^t=\cC[x_i]$ for each $x_i \in \cB$\;
        }
        \tcp{Update model}
        One-step gradient descent on Eq.~\ref{eq:cache-loss}\;
    }
}
\KwOut{Trained model.}
\end{algorithm}

%%%%%%%%%%%%%%%%%%%%%%%%%%%%%%%%%%%%%%%%
\subsection{Memory Saving with KNN}

One caveat of Algorithm~\ref{alg:cache} is the increased memory usage. For example, there are about 4.5 million sentence pairs in the WMT'16 En-De dataset, so that simply caching the adversarial samples takes about 100GB of memory. We propose a memory saving strategy based on K-nearest neighbors (KNN) to address this issue.

\begin{algorithm}[htb!]
\SetAlgoLined
\caption{Memory Saving.}
\label{alg:memory-saving}
\KwIn{$W$: word embedding matrix; $n$: total number of training samples; $p$: proportion of cached samples; $K$: size of each neighbor.}
\tcp{Before training}
Compute $\{v_i\}_{i=1}^n$ using $W$ and Eq.~\ref{eq:sentence-vector}\;
Sample a cache set $\cX \subset \{1\cdots n\}$ such that $|\cX|=\lfloor np \rfloor$\;
\For{$i \in \{1\cdots n\} \setminus \cX$}{
    Find $\cK_i \subset \cX$ for $x_i$ based on cosine similarity among $\{v_i\}_{i=1}^n$\;
}
\tcp{In epoch $t$ where $t \% T_c \neq 0$}
\For{$i=1, \cdots n$}{
    \uIf{$i \in \cX$}{
        Retrieve $\delta^t_i$ from cache\;
    }
    \Else{
        Compute $\delta_i^t$ using $\cK_i$ and Eq.~\ref{eq:construct-adv}\;
    }
}
\end{algorithm}

The idea is to only cache perturbations of some samples, and perturbations of the other samples are constructed using the cached ones on the fly. Specifically, whenever $t\% T_c=0$, i.e., we need to re-compute and re-cache the perturbations, we only cache $\delta_i^t$ such that $i \in \cX$. Here, $\cX \subset \{1 \cdots n\}$ is a pre-defined cache set and $|\cX| \ll n$. This strategy significantly reduces memory overhead.
Consequently, in each epoch $t$ where $t\% T_c \neq 0$, perturbations $\delta_i^t$ such that $i \in \cX$ are directly retrieved from the cache. And perturbations $\delta_i^t$ such that $i \in \{1\cdots n\} \setminus \cX$ are defined as the following:
\begin{align} \label{eq:construct-adv}
    & \delta^t_{i, \ell} = \frac{1}{|\cK_i|} \sum_{j \in \cK_i} \frac{1}{\ell_j} \sum_{\ell'=1}^{\ell_j} \delta^{t-1}_{j,\ell'},
    \ \ell=1, \cdots, \ell_i.
\end{align}
Here, $\ell_i$ be the length of sentence $x_i$, $\delta^t_{i,\ell} \in \RR^d$ is the perturbation for the $\ell$-th word in sentence $x_i$, and $\cK_i$ is the nearest neighbor set for $x_i$ (which we present later).
We remark that constructing the perturbations does not impose extra training time, because we can perform such computation in parallel with training.

We remark that each word has an identical perturbation in Eq.~\ref{eq:construct-adv}, i.e., $\delta^t_i \in \RR^{|\ell_i| \times d}$ has identical rows. We choose this design because a perturbation in the neighbor of $\delta_i^t$ may have a different dimension, i.e., $\delta^t_j \in \RR^{|\ell_j| \times d}$ is in the neighbor of $\delta^t_i$ and it is possible that $|\ell_i| \ne |\ell_j|$. To resolve this issue, we compute the word-level mean of all the perturbations in the neighbor of $\delta_i^t$ and assign it to each row of $\delta^t_i$.

The remaining is to find $K$ nearest neighbors in $\cX$ for each sentence $x_i$ such that $i \in \{1\cdots n\} \setminus \cX$.
Suppose we have a word embedding matrix $W \in \RR^{d \times |\cV|}$, where $|\cV|$ is the vocabulary size and $d$ is the embedding dimension. Note that $W$ can be obtained from pre-trained models such BERT~\cite{devlin2018bert}. For each sentence $x_i$, we compute its sentence representation $v_i \in \RR^d$ as
\begin{align}\label{eq:sentence-vector}
    v_i = \frac{1}{\ell_i} \sum_{\ell=1}^{\ell_i} W x_{i,\ell}.
\end{align}
Here, $x_{i,\ell} \in \RR^{|\cV|}$ is the one-hot vector of the $\ell$-th word in sentence $x_i$. Then, we can find $K$ nearest neighbors $\cK_i$ for sample $x_i$ using the KNN algorithm, where the distance between two samples is defined as their cosine similarity.
Notice that finding $\{\cK_i\}_{i=1}^n$ is a pre-processing step, i.e., we can find the neighbors before training the model.

The memory saving algorithm is summarized in Algorithm~\ref{alg:memory-saving}, and an extended version that combines caching and memory saving is presented in Algorithm~\ref{alg:full} in the appendix.

%%%%%%%%%%%%%%%%%%%%%%%%%%%%%%%%%%%%%%%%
\subsection{Computational Efficiency}

Computational costs of various methods are summarized in Table~\ref{tb:comp-cost}. In conventional adversarial regularization algorithms, such as FreeLB~\cite{zhu2019freelb} and SMART~\cite{jiang2019smart}, suppose we solve the inner maximization problem for $S$ steps, then we impose extra $S$ forward passes and $S$ backward passes in each iteration.
In contrast, R3F~\cite{aghajanyan2020better} removes the maximization problem, and directly samples perturbations from a normal distribution. Thus, R3F only introduce one extra forward pass to compute the regularization term.
Using Algorithm~\ref{alg:cache}, our method shares similar efficiency as R3F. Specifically, suppose we cache the perturbations every $T_c$ epochs, then the average number of forward passes and backward passes per iteration is $2+(S-1)/T_c$ and $1+S/T_c$, respectively. In practice, $S/T_c$ is usually small, such that the computational cost between \ours and R3F is close.

\begin{figure}[tb!]
    \centering
    \includegraphics[width=0.49\linewidth]{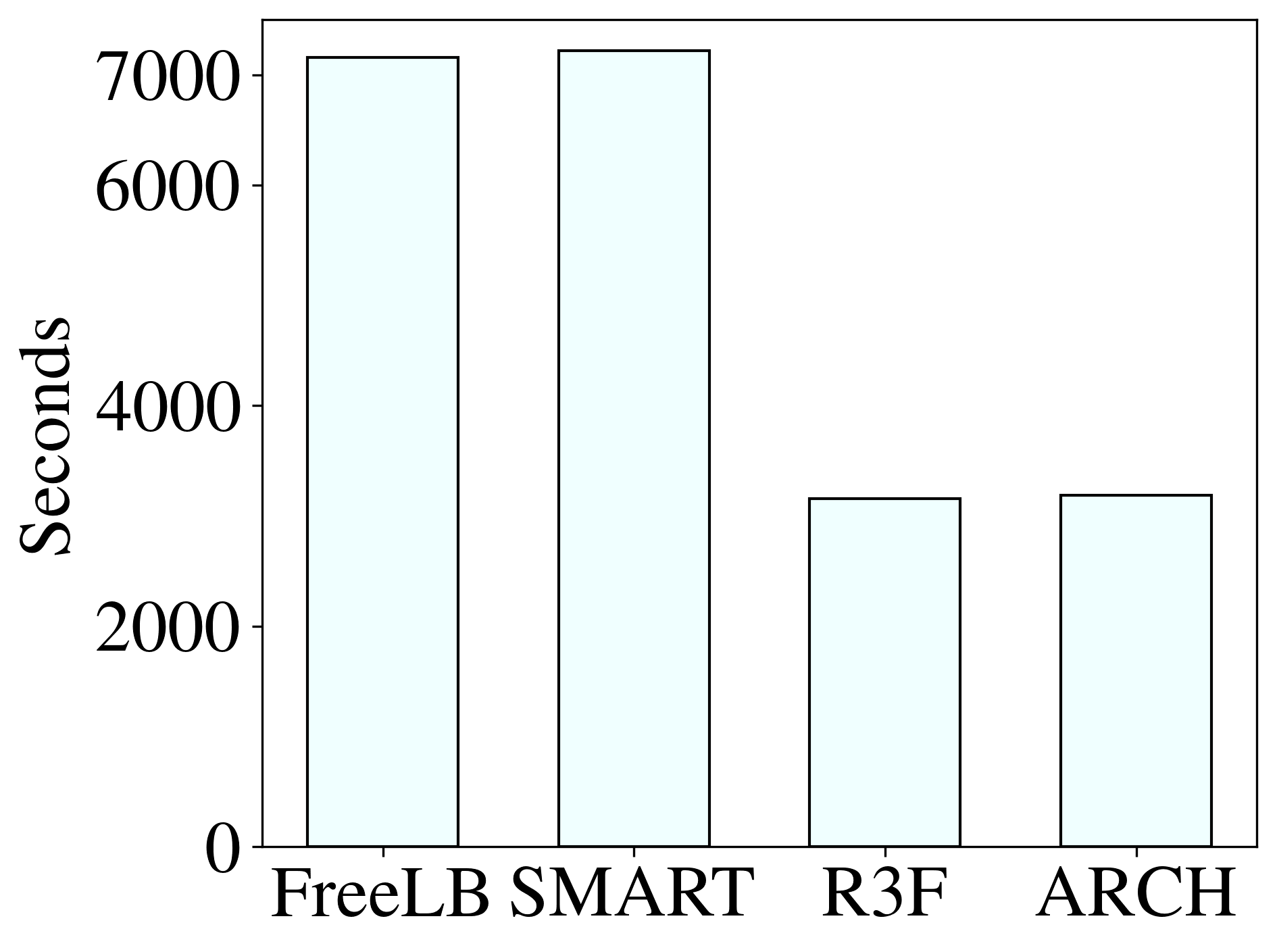}
    \includegraphics[width=0.49\linewidth]{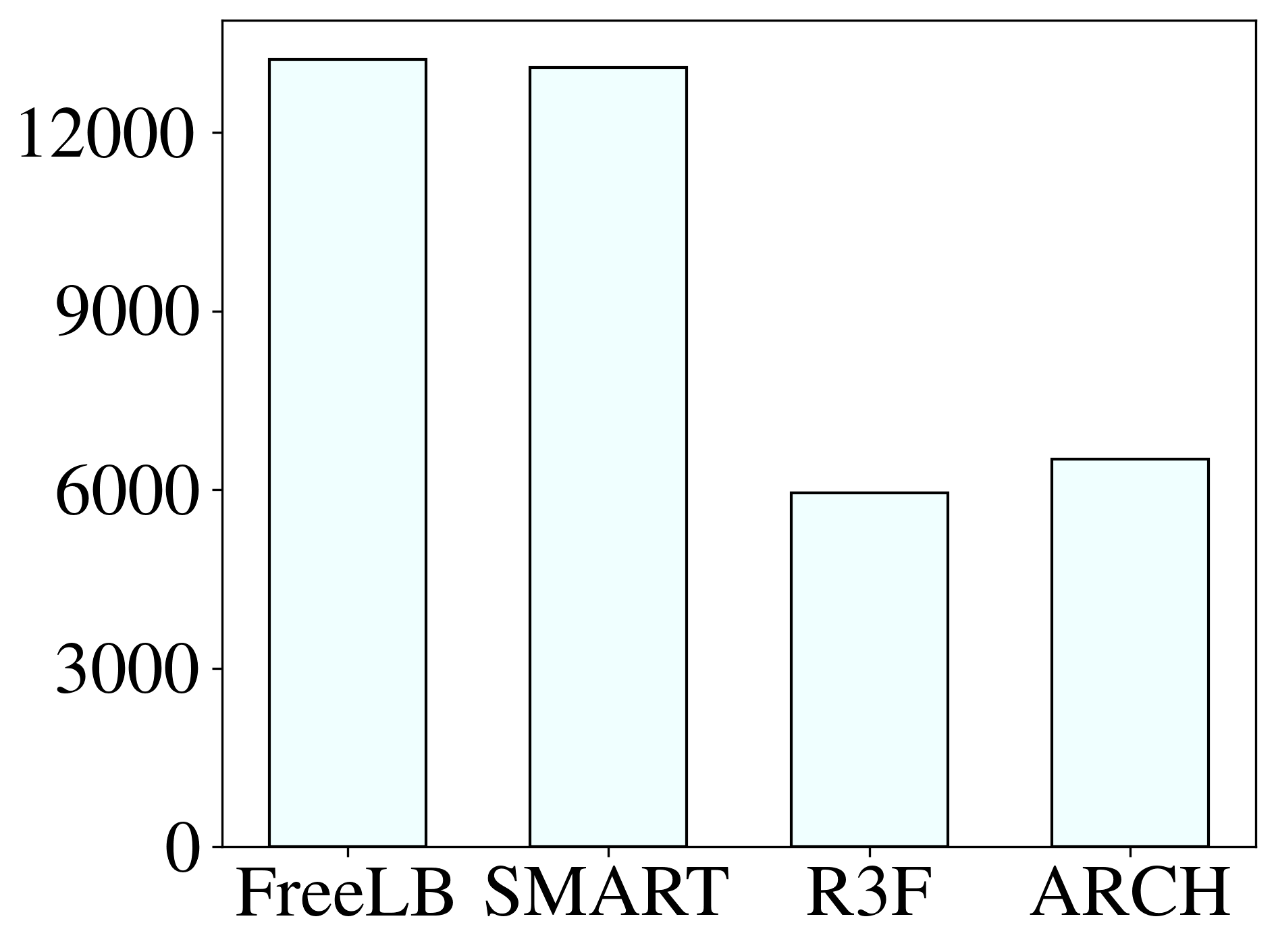}
    \vskip -0.05in
    \caption{Wall time of different methods. Left: training a Transformer-base model for 150 epochs on IWSLT'14 De-En; Right: fine-tuning BERT\textsubscript{BASE} for 10 epochs on SST-2.}
    \label{fig:wall-time}
    % \vskip -0.1in
\end{figure}

Wall time comparison is illustrated in Fig.~\ref{fig:wall-time}. Notice that in the left subfigure, both our method and R3F save about 70\% computation time in comparison with FreeLB and SMART. In the right subfigure, the time saving is about 50\%.
The absolute time saving is more significant on large models and large datasets. For example, when training a Transformer-big model on the WMT'16 En-De dataset, our method costs about 176 GPU hours, while SMART uses 576 GPU hours. 

\begin{table}[tb!]
\centering
\begin{tabular}{l|cc}
\toprule
& \textbf{Forward} & \textbf{Backward} \\ \midrule
Standard & $1$ & $1$ \\
FreeLB & $1+S$ & $1+S$ \\
SMART & $1+S$ & $1+S$ \\
R3F & $2$ & $1$ \\
\ours & $2+(S-1)/T_c$ & $1+S/T_c$ \\ \bottomrule
\end{tabular}
\vskip -0.05in
\caption{Computational cost of various methods. Here \textit{S} is the number of gradient ascent (PGD) steps, and $T_c$ is the number of epochs between caching. \textit{Forward} is the number of forward passes, and \textit{Backward} is the number of backward passes.}
\label{tb:comp-cost}
% \vskip -0.05in
\end{table}

%% file: 0-experiments.tex
\newcolumntype{C}{@{\hskip3pt}c@{\hskip3pt}}
\begin{table*}[!t]
\centering
\begin{tabular}{l|cccccc}
\toprule
Models & \textbf{En-Vi} & \textbf{Vi-En} & \textbf{En-De} & \textbf{De-En} & \textbf{En-Fr} & \textbf{Fr-En} \\ \midrule
Transformer~\cite{vaswani2017attention} & 30.3 & 28.7 & 28.3 & 34.7 & 39.3 & 38.2 \\ 
R3F~\cite{aghajanyan2020better} & 31.6 & 30.0 & 29.0 & 35.4 & 39.5 & 38.7 \\
FreeLB~\cite{zhu2019freelb} & 31.6 & 29.6 & 28.6 & 35.3 & 39.4 & 38.7 \\ 
SMART~\cite{jiang2019smart} & 31.5 & 30.1 & 29.2 & 35.5 & 39.8 & 38.9 \\ \midrule 
\ours & \textbf{32.0} & \textbf{30.4} & \textbf{29.4} & \textbf{36.1} & \textbf{40.3} & \textbf{39.3} \\
\bottomrule
\end{tabular}
\vskip -0.05in
\caption{BLEU score on three low-resource datasets. All the baseline results are from our re-implementation. We report the mean over three runs using different random seeds. \textbf{\ours saves about 70\% computational time comparing with SMART.}}
\label{tb:low-resource}
% \vskip -0.1in
\end{table*}

\begin{table}[!t]
\centering
\begin{tabular}{l|cc}
\toprule
Models & \textbf{BLEU} & \textbf{sacreBLEU} \\ \midrule 
Transformer & 29.1 & 28.4 \\
R3F & 29.4 & 29.0 \\
FreeLB & 29.3 & 29.0 \\ 
SMART & \textbf{29.8} & 29.1 \\ \midrule 
\ours & \textbf{29.8} & \textbf{29.4} \\ 
\bottomrule
\end{tabular}
\vskip -0.05in
\caption{BLEU and sacreBLEU score on the WMT'16 En-De dataset. All the baseline results are from our re-implementation.}
\label{tb:rich-resource}
\end{table}

\begin{table}[!t]
\centering
\begin{tabular}{c|cccc}
\toprule 
\textbf{Data}  & \textbf{Source} & \textbf{Train} & \textbf{Valid} & \textbf{Test} \\ \midrule
\textbf{En-Vi} & IWSLT'15        & 133k           & 768            & 1268          \\
\textbf{En-De} & IWSLT'14        & 161k           & 7.2k           & 6.7k          \\
\textbf{En-Fr} & IWSLT'16        & 224k           & 1080           & 1133          \\
\textbf{En-De} & WMT'16          & 4.5m           & 3.0k           & 3.0k          \\ 
\bottomrule
\end{tabular}
\vskip -0.05in
\caption{Dataset source and statistics. Here ``k'' stands for thousand, and ``m'' stands for million.}
\label{tb:dataset}
\end{table}

%%%%%%%%%%%%%%%%%%%%%%%%%%%%%%%%%%%%%%%%%%%%%%%%%%%%%%%%%%%%
\section{Experiments}

In all the experiments, we use \textit{PyTorch}\footnote{\url{https://pytorch.org/}} \cite{paszke2019pytorch} as the backend. All the experiments are conducted on NVIDIA V100 GPUs.

%%%%%%%%%%%%%%%%%%%%%%%%%%%%%%%%%%%%%%%%%%%%%%%%%%%%%%%%%%%%
\subsection{Baselines}

We adopt several baselines in the experiments.

\vspace{0.05in}
\noindent $\diamond$ \textit{Transformer} \cite{vaswani2017attention} achieves superior performance in neural machine translation.

\vspace{0.05in}
\noindent $\diamond$ \textit{BERT} \cite{devlin2018bert} exhibits outstanding performance when fine-tuned on natural language understanding tasks.

\vspace{0.05in}
\noindent $\diamond$ \textit{FreeAT} \cite{shafahi2019adversarial} enables ``free'' adversarial training by recycling the gradient information generated when updating the model.

\vspace{0.05in}
\noindent $\diamond$ \textit{FreeLB} \cite{zhu2019freelb} treats the intermediate perturbations during the projected gradient ascent steps as virtual batches. As such, the method achieves ``free'' large batch adversarial training. 

\vspace{0.05in}
\noindent $\diamond$ \textit{SMART} \cite{jiang2019smart} achieves state-of-the-art performance in natural language understanding. The method utilizes smoothness-inducing regularization and Bregman proximal point optimization.

\vspace{0.05in}
\noindent $\diamond$ \textit{R3F} \cite{aghajanyan2020better} replaces the maximization problem in conventional adversarial regularization with random noise.

\begin{table*}[t!]
\centering
\resizebox{1.0\textwidth}{!}{
\begin{tabular}{@{\hskip3pt}l@{\hskip3pt}|C|C|C|C|C|C|C|C|C}
\bottomrule
& \textbf{RTE} & \textbf{MRPC} & \textbf{CoLA} & \textbf{SST-2} & \textbf{STS-B} & \textbf{QNLI} & \textbf{QQP} & \textbf{MNLI-m/mm} & \textbf{Average} \\
& Acc & Acc/F1 & Mcc & Acc & P/S Corr & Acc & Acc/F1 & Acc & \textbf{Score} \\ \midrule
BERT\textsubscript{BASE} & 63.5 & 84.1/89.0 & 54.7 & 92.9 & 89.2/88.8 & 91.1 & 90.9/88.3 & 84.5/84.4 & 81.5 \\
FreeAT & 68.0 & 85.0/89.2 & 57.5 & 93.2 & 89.5/89.0 & 91.3 & 91.2/88.5 & 84.9/85.0 & 82.6 \\
FreeLB & 70.0 & 86.0/90.0 & 58.9 & 93.4 & 89.7/89.2 & 91.5 &91.4/88.4& 85.4/85.5 & 83.3 \\
R3F & 70.4 & 87.0/91.0 & 59.1 & 93.4 & 90.1/89.8 & 92.0 & 91.7/88.8 & 85.2/85.4 & 83.7 \\
SMART & 71.2 & 87.7/91.3 & 59.1 & 93.0 & 90.0/89.4 & 91.7& 91.5/88.5 & \textbf{85.6/86.0} & 83.8 \\ \hline
\ours & \textbf{72.2} & \textbf{88.0/91.6} & \textbf{61.1} & \textbf{93.6} & \textbf{90.6/90.2} & \textbf{92.2} &\textbf{91.9/89.1} & \textbf{85.6/86.0} & \textbf{84.5} \\
\bottomrule
\end{tabular}
}
\vskip -0.05in
\caption{Evaluation results on the GLUE development set. We use the \textit{BERT\textsubscript{BASE}} architecture for all the methods. The best results on each dataset are shown in \textbf{bold}. Results of \textit{BERT\textsubscript{BASE}}~\cite{devlin2018bert}, \textit{FreeAT}~\cite{shafahi2019adversarial}, \textit{FreeLB}~\cite{zhu2019freelb}, and \textit{R3F}~\cite{aghajanyan2020better} are based on our re-implementation. \textit{SMART} results are from \citet{jiang2019smart}.}
\label{tb:glue-results}
% \vskip -0.1in
\end{table*}

% \begin{table*}[htb!]
% \centering
% \begin{tabular}{@{\hskip3pt}l@{\hskip3pt}|C|C|C|C|C|C|C|C|C}
% \bottomrule
% & \textbf{RTE} & \textbf{MRPC} & \textbf{CoLA} & \textbf{SST-2} & \textbf{STS-B} & \textbf{QNLI} & \textbf{QQP} & \textbf{MNLI-m/mm} & \textbf{Average} \\
% & Acc & Acc/F1 & Mcc & Acc & P/S Corr & Acc & Acc/F1 & Acc & \textbf{Score} \\ \midrule
% BERT\textsubscript{BASE} & 66.4 & 84.8/88.9 & 52.1 & 93.5 & 87.1/85.8 & 90.5 & 71.2/89.2 & 84.6/83.4 & 80.0 \\
% FreeLB & 70.1 & 83.5/88.1 & 54.5 & 93.6 & 87.7/86.7 & 91.8 & 72.7/89.6 & 85.7/84.6 & 81.2 \\ \hline
% \ours & \textbf{70.4} & \textbf{85.0/89.2} & \textbf{55.4} & \textbf{94.2} & \textbf{87.9/86.9} & \textbf{91.9} & 72.3/89.5 & 85.6/\textbf{84.7} & \textbf{81.6} \\
% \bottomrule
% \end{tabular}
% \vskip -0.1in
% \caption{GLUE test set results on the GLUE evaluation server. All the methods fine-tune a pre-trained \textit{BERT\textsubscript{BASE}} model. \textit{FreeAT}, \textit{R3F} and \textit{SMART} did not report BERT\textsubscript{BASE} results in their paper or on the GLUE evaluation server. Model references: \textit{BERT\textsubscript{BASE}} \cite{devlin2018bert}, \textit{FreeLB} \cite{zhu2019freelb}.}
% \label{tb:glue-test-results}
% \vskip -0.1in
% \end{table*}

%%%%%%%%%%%%%%%%%%%%%%%%%%%%%%%%%%%%%%%%
\subsection{Machine Translation}

%%%%%%%%%%%%%%%%%%%%%%%%%%%%%%
\textbf{Datasets.}
We use three low-resource datasets\footnote{\url{https://iwslt.org/}}: English-German from IWSLT'14, English-Vietnamese from IWSLT'15, and English-French from IWSLT'16. We also use a rich-resource dataset: English-German from WMT'16.
Dataset statistics are summarized in Table~\ref{tb:dataset}.

%%%%%%%%%%%%%%%%%%%%%%%%%%%%%%
\vspace{0.05in}
\noindent
\textbf{Implementation.}
In NMT tasks, we have the source-side and the target-side inputs. We add perturbations to both of their embeddings~\cite{sano2019effective}. This has demonstrated to be more effective than adding perturbations to a single side.
We use \textit{Fairseq}\footnote{\url{https://github.com/pytorch/fairseq}} \cite{ott2019fairseq} to implement our algorithms.
For En-Vi and En-Fr experiments, we use the Transformer-base architecture~\cite{vaswani2017attention}. For En-De (IWSLT'14) experiments, we modify\footnote{\url{https://github.com/pytorch/fairseq/tree/master/examples/translation}} the Transformer-base architecture by decreasing the hidden dimension size from 2048 to 1024, and decreasing the number of heads from 8 to 4 (while dimension of each head doubles). For En-De (WMT'16) experiments, we use the Transformer-big~\cite{vaswani2017attention} architecture. The training details are presented in Appendix~\ref{app:nmt}.

%%%%%%%%%%%%%%%%%%%%%%%%%%%%%%
\vspace{0.05in}
\noindent
\textbf{Results.}
Experimental results on the low-resource datasets are summarized in Table~\ref{tb:low-resource}. We can see that \ours outperforms all the baselines in all the experiments. We remark that our method saves about 70\% computational time in comparison with SMART and FreeLB, and has the save level of efficiency comparing with R3F (Fig.~\ref{fig:wall-time}).
Even though R3F is efficient by eliminating the maximization problem, we can see that is does not generalize as well as SMART, i.e., R3F has worse BLEU score than SMART in 5/6 of the experiments.

Experimental results on the WMT'16 En-De dataset are summarized in Table~\ref{tb:rich-resource}. We report both the BLEU score and the sacreBLEU~\cite{post2018call} score. The former is standard for machine translation tasks, and the latter is a detokenzied version of BLEU.
The absolute computational time saving is more significant for larger datasets (e.g., WMT) and larger models (e.g., Transformer-big). In the experiments, \ours uses about 176 GPU hours to train, while it costs SMART about 576 hours.
Performance of \ours is better or on par with all the baselines. Notice that like in Table~\ref{tb:low-resource}, performance of R3F is worse than SMART.

%%%%%%%%%%%%%%%%%%%%%%%%%%%%%%%%%%%%%%%%
\subsection{Natural Language Understanding}

%%%%%%%%%%%%%%%%%%%%%%%%%%%%%%
\noindent
\textbf{Datasets.}
We conduct experiments on the General Language Understanding Evaluation (GLUE) benchmark \cite{wang2018glue}, which is a collection of nine natural language inference tasks. The benchmark includes question answering~\cite{squad1}, linguistic acceptability (CoLA, \citealt{cola2018}), sentiment analysis (SST, \citealt{sst2013}), text similarity (STS-B, \citealt{sts-b2017}), paraphrase detection (MRPC, \citealt{mrpc2005}), and natural language inference (RTE \& MNLI, \citealt{rte1,rte2,rte3,rte5,mnli2018}) tasks. Statistics of the datasets are summarized in Table~\ref{tb:glue} (Appendix~\ref{app:glue}).

%%%%%%%%%%%%%%%%%%%%%%%%%%%%%%
\vspace{0.05in}
\noindent
\textbf{Implementation.}
We implement our algorithm using the \textit{MT-DNN}\footnote{\url{https://github.com/namisan/mt-dnn}}~\cite{liu2019multi, mtdnn2020demo} and the Transformers~\cite{wolf-etal-2020-transformers} code-base. The training details are presented in Appendix~\ref{app:glue}.

%%%%%%%%%%%%%%%%%%%%%%%%%%%%%%
\vspace{0.05in}
\noindent
\textbf{Results.}
Table~\ref{tb:glue-results} summarizes experimental results on the GLUE development set. We can see that \ours is on par or outperforms all the baselines in all the tasks. Notice that generalization of R3F is comparable with SMART. Our proposed method shares the advantages of both efficiency (i.e., R3F) and informative perturbations (i.e., SMART), and thus, \ours behaves better than both of these methods.
We highlight that our method is 50\%-70\% faster than SMART and FreeLB.

%%%%%%%%%%%%%%%%%%%%%%%%%%%%%%%%%%%%%%%%
\subsection{Parameter Study}

\begin{figure}[!t]
    \centering
    \begin{subfigure}{0.24\textwidth}
        \centering
        \includegraphics[width=1.0\textwidth]{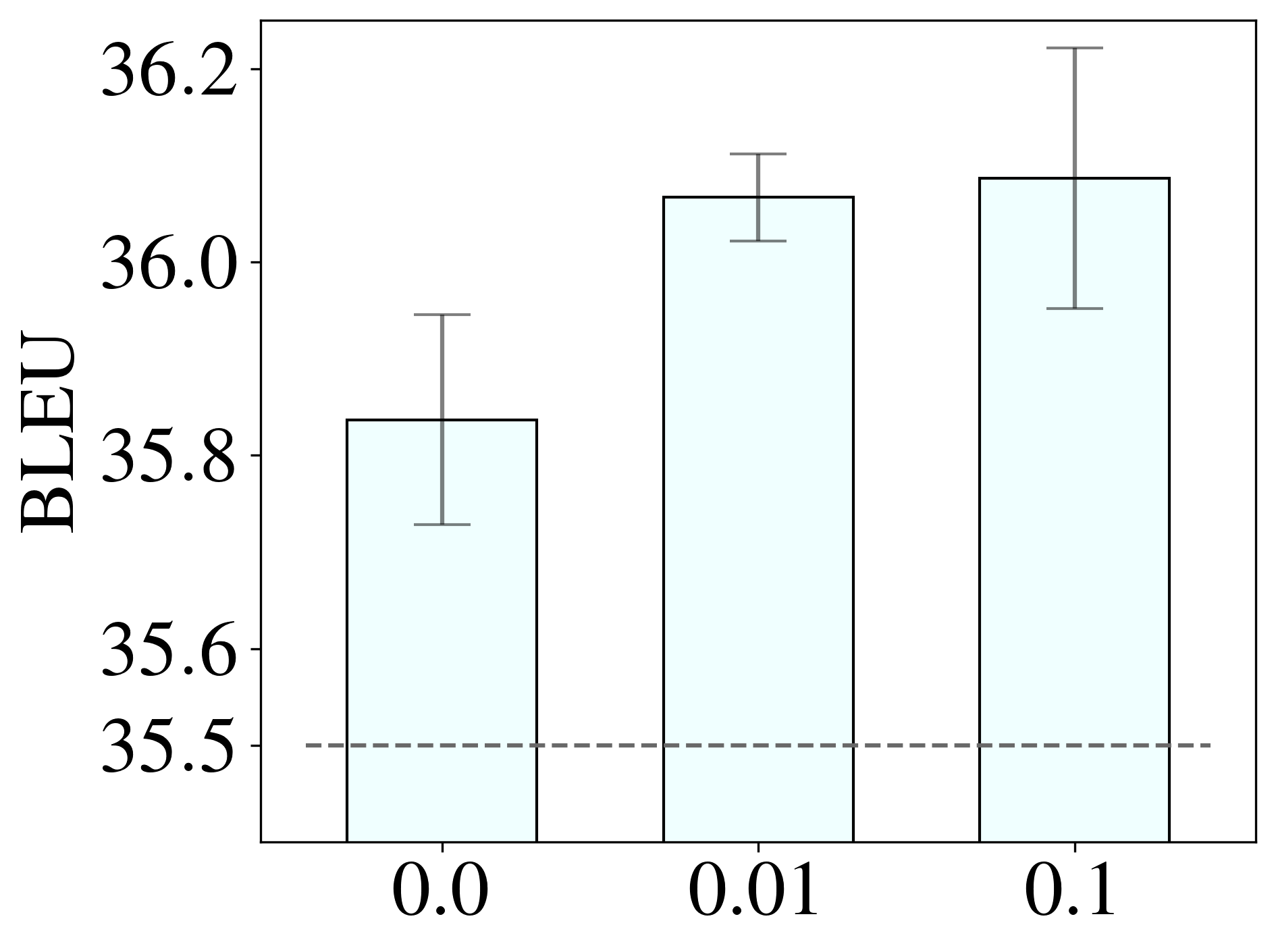}
        % \vskip -0.05in
        \caption{Moving average.}
        \label{fig:param:alpha}
    \end{subfigure}%
    \begin{subfigure}{0.24\textwidth}
        \centering
        \includegraphics[width=1.0\linewidth]{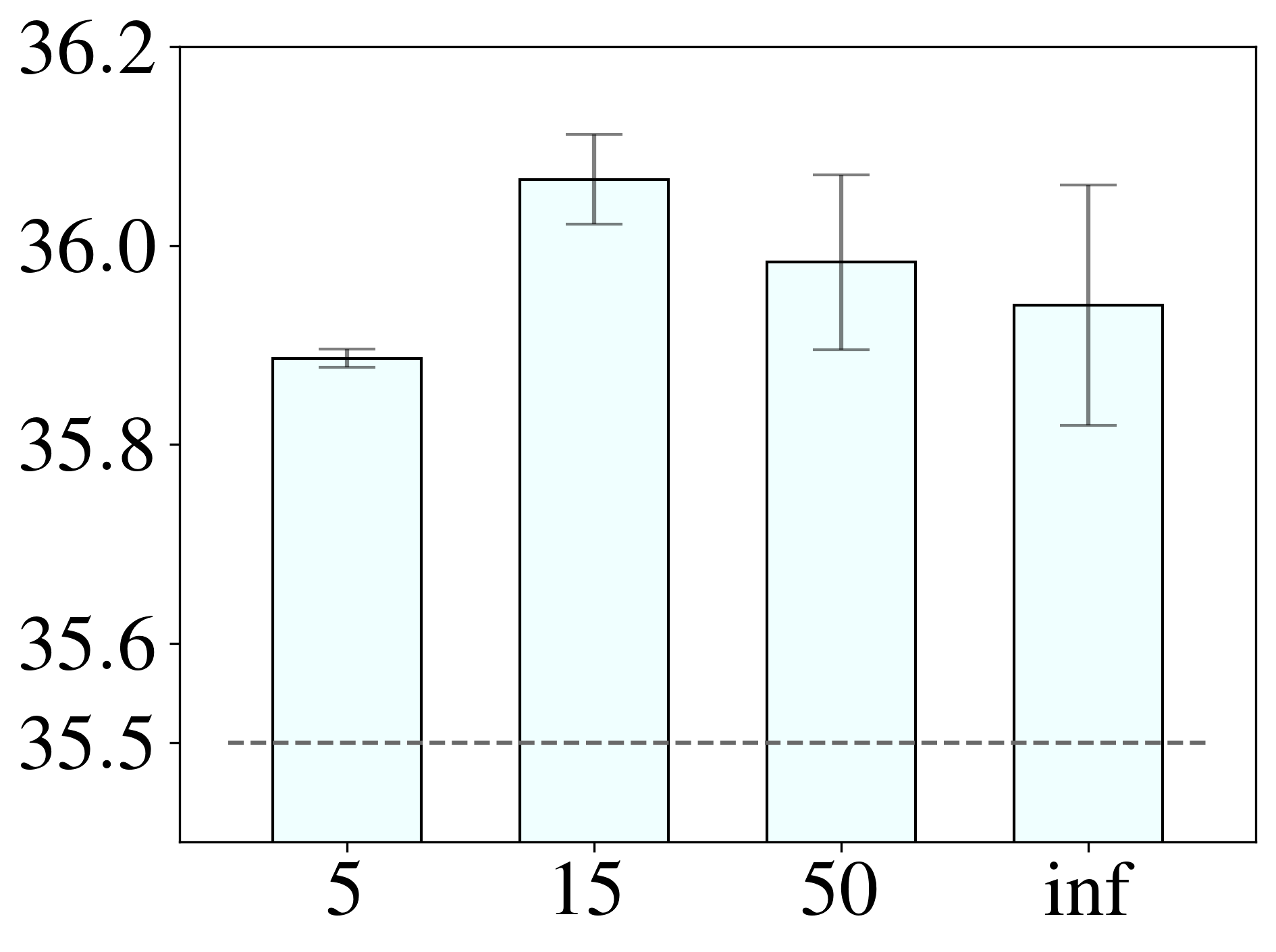}
        % \vskip -0.05in
        \caption{Epochs between caching.}
        \label{fig:param:cache_every}
    \end{subfigure}
    \vskip 0.05in
    \begin{subfigure}{0.24\textwidth}
        \centering
        \includegraphics[width=1.0\linewidth]{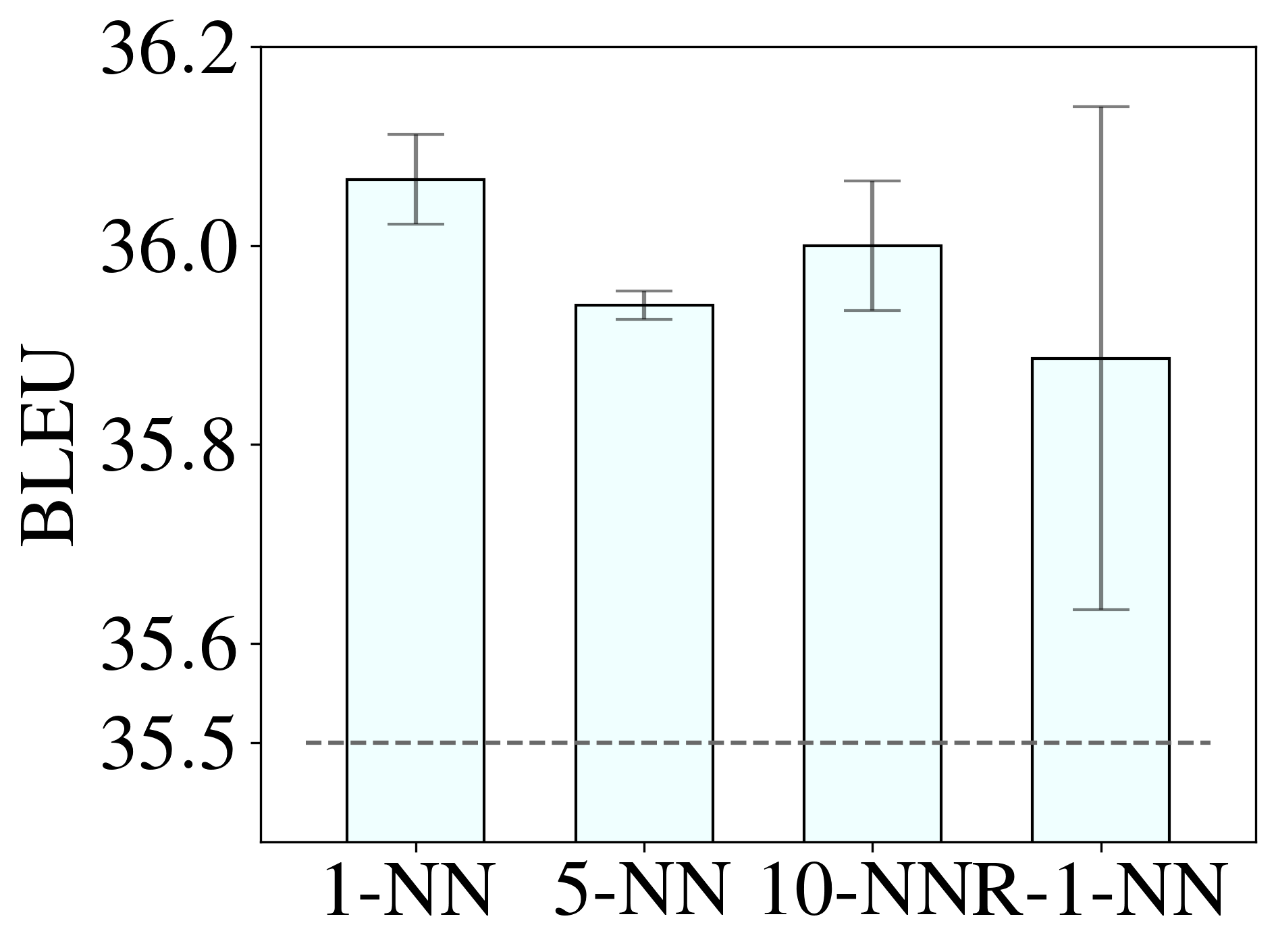}
        % \vskip -0.05in
        \caption{Number of neighbors.}
        \label{fig:param:neighbor}
    \end{subfigure}%
    \begin{subfigure}{0.24\textwidth}
        \centering
        \includegraphics[width=1.0\linewidth]{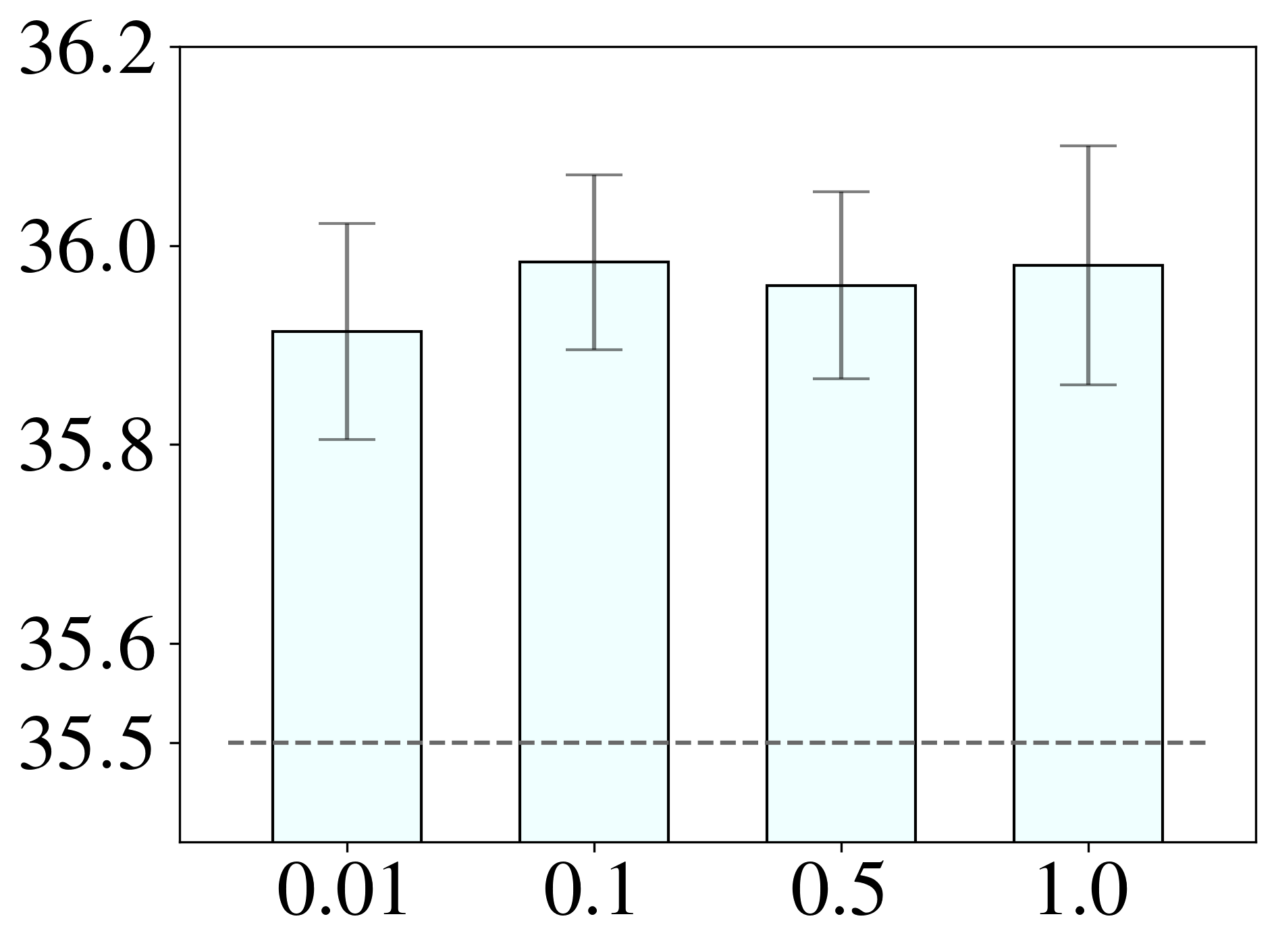}
        % \vskip -0.05in
        \caption{Proportion of caching.}
        \label{fig:param:p_cache}
    \end{subfigure}%
    \vskip -0.05in
    \caption{Parameter study on the IWSLT'14 De-En dataset. Each error bar is based on three runs using different random seeds. Each dashed line signifies the SMART baseline. In (b), \textit{inf} means we only cache once during training. In (c), \textit{R-1-NN} means we use 1-NN, but the neighbor is randomly selected.}
    \label{fig:analysis}
    % \vskip -0.1in
\end{figure}

\noindent $\diamond$
\textbf{Moving average helps.}
As indicated in Fig.~\ref{fig:param:alpha}, without the exponential moving average, model performance drops about 0.3 BLEU. Also, the model is robust to the moving average parameter, as increasing it from 0.01 to 0.1 does not change model performance.

\vspace{0.05in}
\noindent $\diamond$
\textbf{Number of epochs between caching is important.}
If we cache the perturbations too frequently (i.e., 5 in Fig.~\ref{fig:param:cache_every}), the model cannot adapt to the perturbations well; and if we cache the perturbations too infrequently (i.e., inf in Fig.~\ref{fig:param:cache_every}), staleness of the perturbations hinders model generalization.

\vspace{0.05in}
\noindent $\diamond$
\textbf{Robustness to the number of neighbors.}
In Fig.~\ref{fig:param:neighbor}, notice that \ours is robust to the number of neighbors. We also examine a variant of the KNN memory-saving strategy (R-1-NN): namely in Algorithm~\ref{alg:cache}, the nearest neighbors set $\cK_i$ for sample $x_i$ is randomly constructed instead of based on word embeddings. We can see that model performance drops, and the method also exhibits drastically larger variance.

\vspace{0.05in}
\noindent $\diamond$
\textbf{Robustness to the number of cached samples.}
From Fig.~\ref{fig:param:p_cache}, notice that the model generalizes well even caching only 1\% of the perturbations (i.e., only 1400 samples for the IWSLT'14 De-En dataset). Moreover, the KNN memory-saving strategy does not hinder model performance, i.e., the BLEU score is consistent when caching all the samples and caching only 10\% of the samples. 

We highlight that in practice \textbf{\ours does not need much tuning}, because the method is robust to the introduced hyper-parameters.

%%%%%%%%%%%%%%%%%%%%%%%%%%%%%%%%%%%%%%%%
\subsection{Analysis}

\begin{figure}[htb!]
    \centering
    \includegraphics[width=0.8\linewidth]{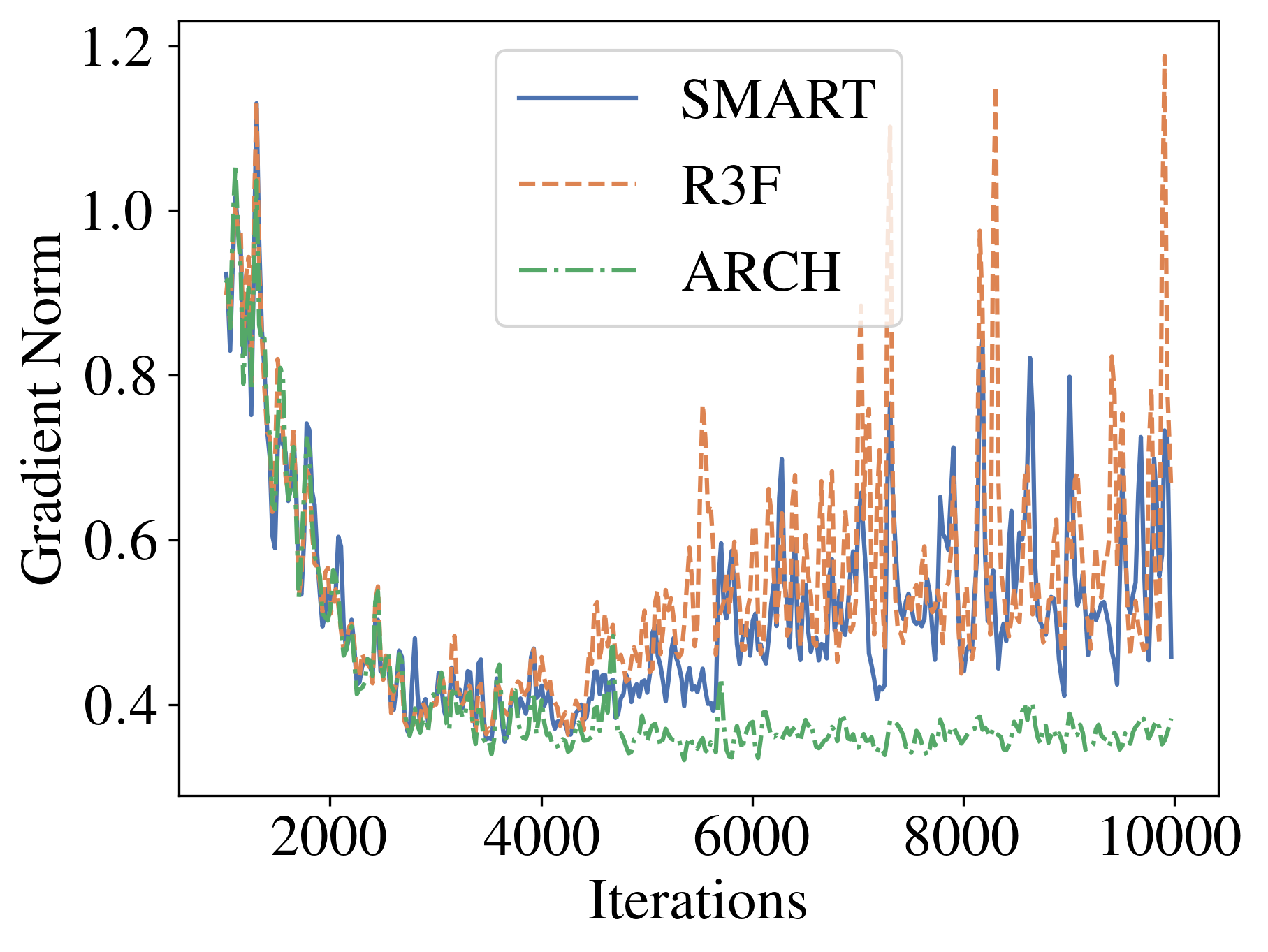}
    \includegraphics[width=0.8\linewidth]{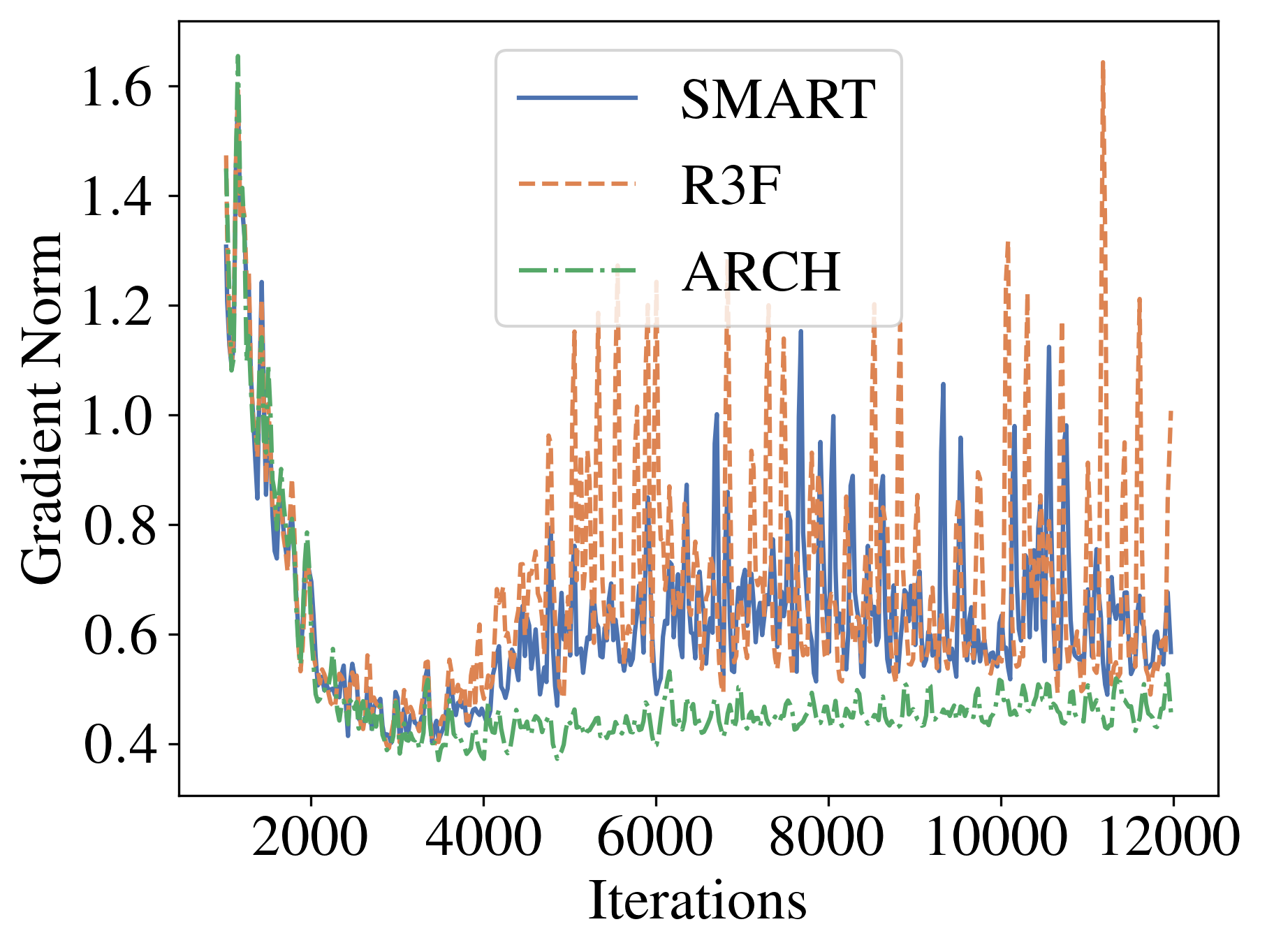}
    \vskip -0.05in
    \caption{Norm of stochastic gradients during training. Top: IWSLT'14 De-En; Bottom: IWSLT'15 En-Vi.}
    \label{fig:grad-norm}
    % \vskip -0.15in
\end{figure}

\noindent $\diamond$
\textbf{Caching reduces gradient norm variance.}
As demonstrated in Fig.~\ref{fig:grad-norm}, variance of the gradient norms reduces significantly comparing with SMART and R3F. This meets our expectation that by reusing perturbations, the model can adapt to the noisy data (i.e., clean data with perturbations) better. Notice that R3F has even larger gradient norm variance than SMART, which is because R3F uses random noise instead of data-dependent ones.

%%%%%%%%%%%%%%%%%%%%%%%%%%%%%%
\vspace{0.05in}
\noindent $\diamond$
\textbf{Adversarial robustness.}
We remark that the focus of \ours is model generalization. Nevertheless, we investigate model robustness on the Adversarial-NLI (ANLI, \citealt{nie2019adversarial}) dataset. The dataset contains 163k data, which are collected via a human-and-model-in-the-loop approach. Surprisingly, from Table~\ref{tb:anli}, we can see that R3F and \ours achieve on par robustness with SMART. This indicates that reusing perturbations, or even constructing random perturbations can increase robustness (than BERT) to the same level as computing optimized perturbations (i.e., SMART).

\begin{table}[htb!]
\centering
\begin{tabular}{l|cccc}
\toprule
& \multicolumn{4}{c}{Dev} \\
& R1 & R2 & R3 & All \\ \midrule
BERT\textsubscript{BASE} & 53.3 & 43.0 & 44.7 & 46.8 \\
% FreeLB & 53.4 & 44.8 & 44.9 & 47.5 \\
R3F & 53.9 & 43.4 & \textbf{46.3} & 47.8 \\
SMART & \textbf{54.1} & 44.4 & 45.3 & 47.8 \\ \hline
\ours & 54.0 & \textbf{46.1} & 46.0 & \textbf{48.5} \\ \bottomrule
& \multicolumn{4}{c}{Test} \\
& R1 & R2 & R3 & All \\ \midrule
BERT\textsubscript{BASE} & 54.1 & 44.9 & 46.6 & 48.4 \\
% FreeLB & 53.5 & 44.7 & \textbf{47.6} & 48.5 \\
R3F & \textbf{54.3} & 46.2 & 46.5 & 48.8 \\
SMART & \textbf{54.3} & 46.4 & 46.5 & 48.9 \\ \hline
\ours & 53.8 & \textbf{46.6} & \textbf{47.4} & \textbf{49.2} \\ \bottomrule
\end{tabular}
\vskip -0.05in
\caption{Experimental results on the ANLI dataset. Model references: \textit{BERT\textsubscript{BASE}} \cite{devlin2018bert}, R3F \cite{aghajanyan2020better}, \textit{SMART} \cite{jiang2019smart}.}
\label{tb:anli}
% \vspace{-0.1in}
\end{table}

\begin{figure}[htb!]
    \centering
    \includegraphics[width=0.48\linewidth]{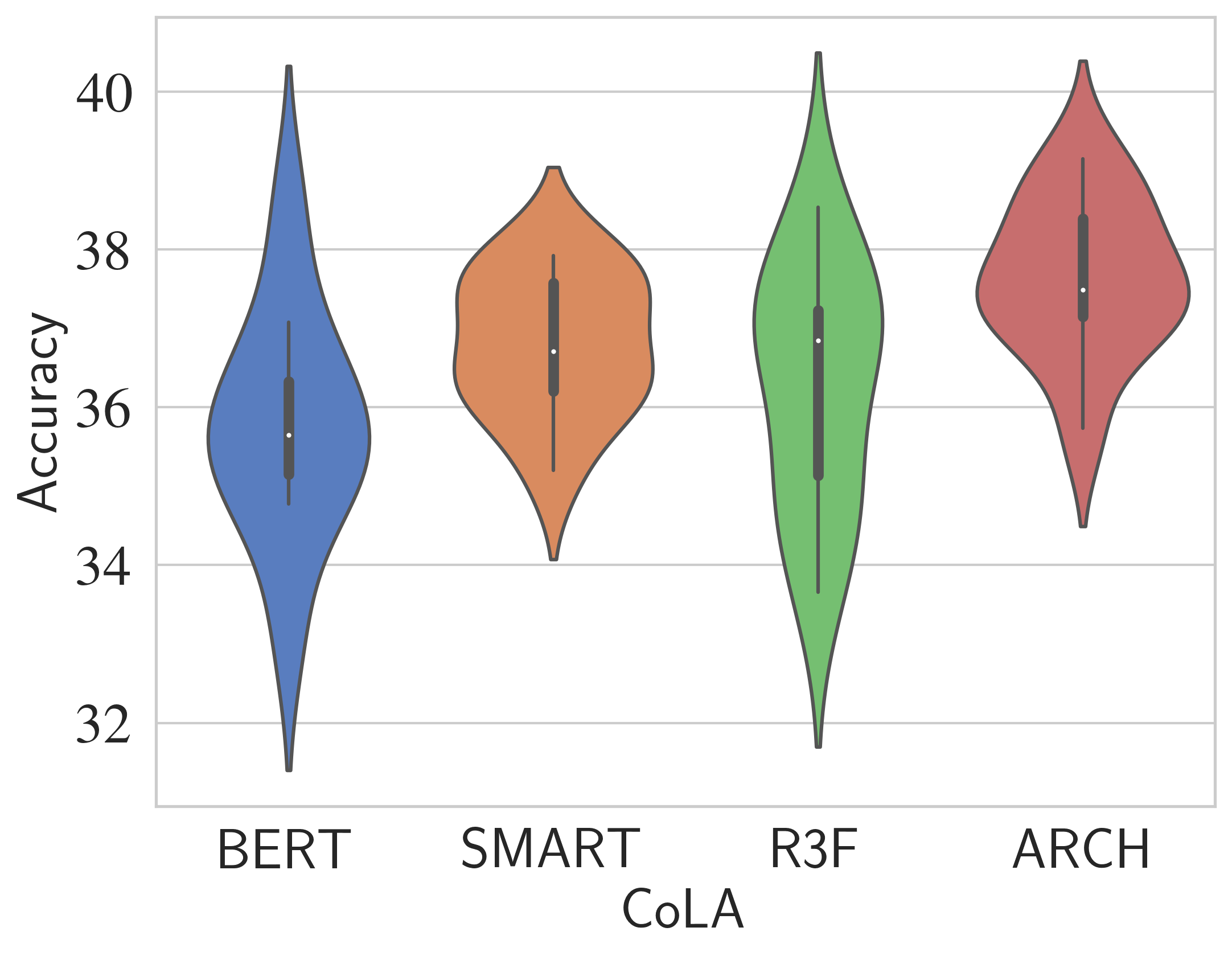}
    \includegraphics[width=0.48\linewidth]{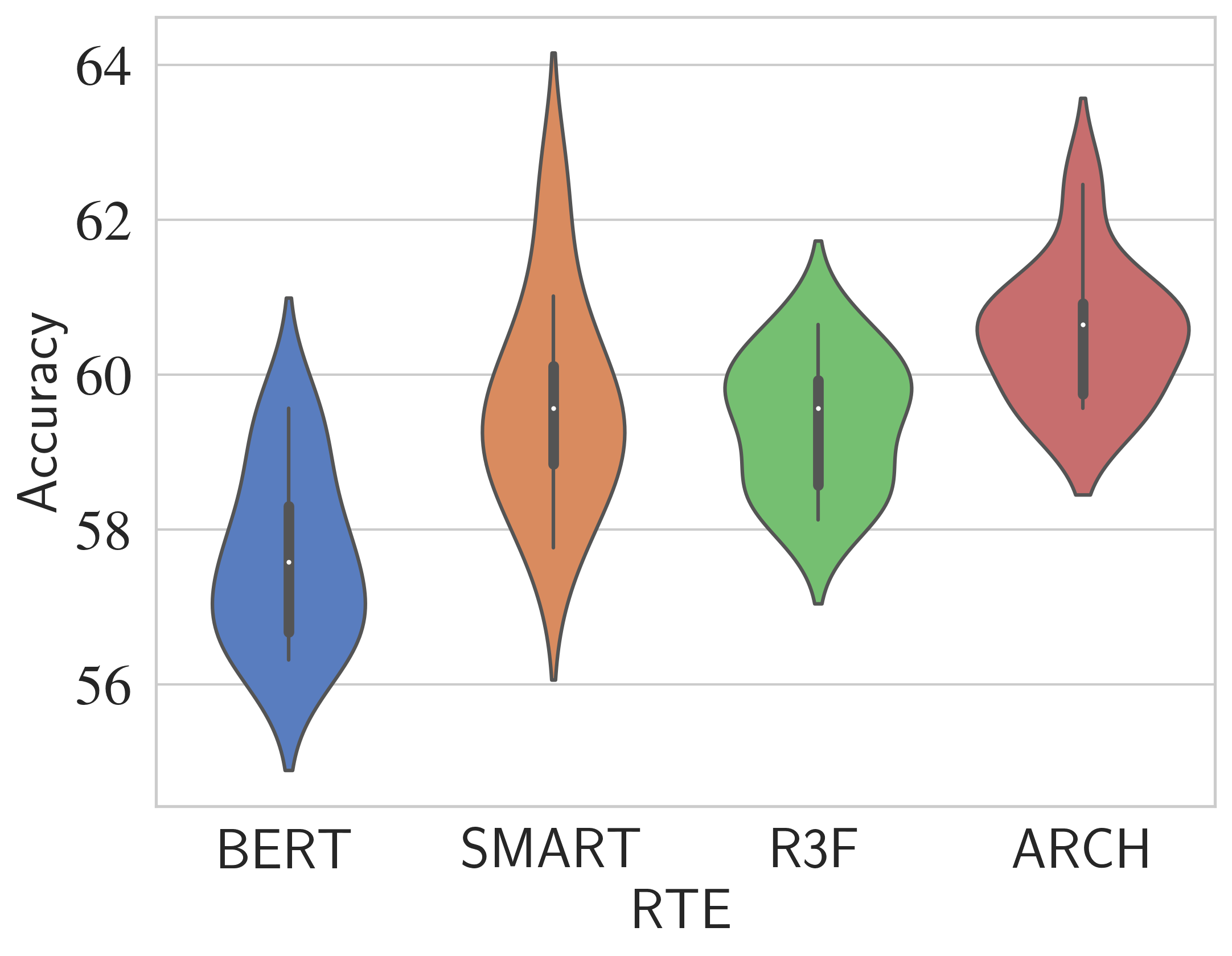}
    \includegraphics[width=0.48\linewidth]{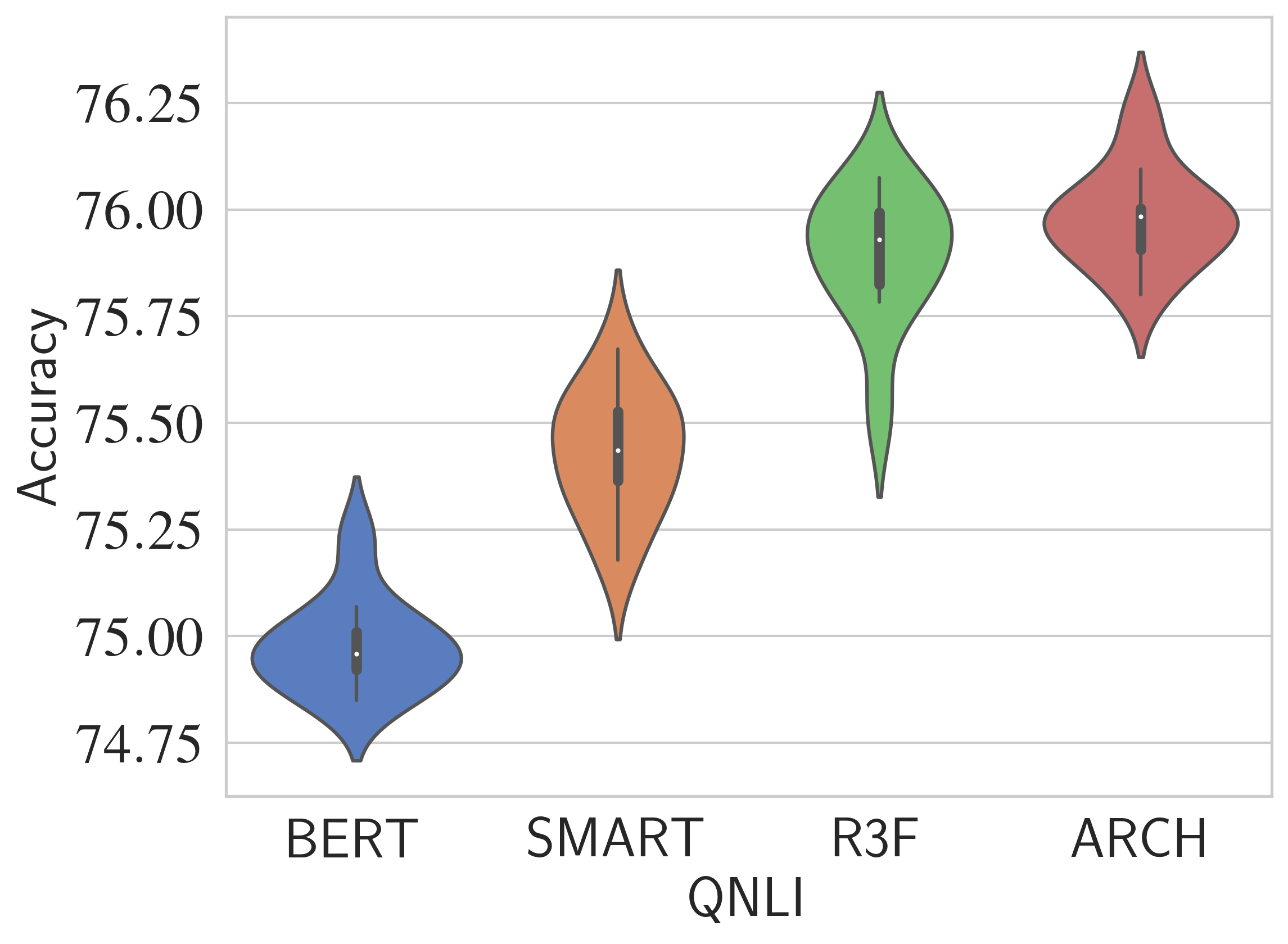}
    \includegraphics[width=0.48\linewidth]{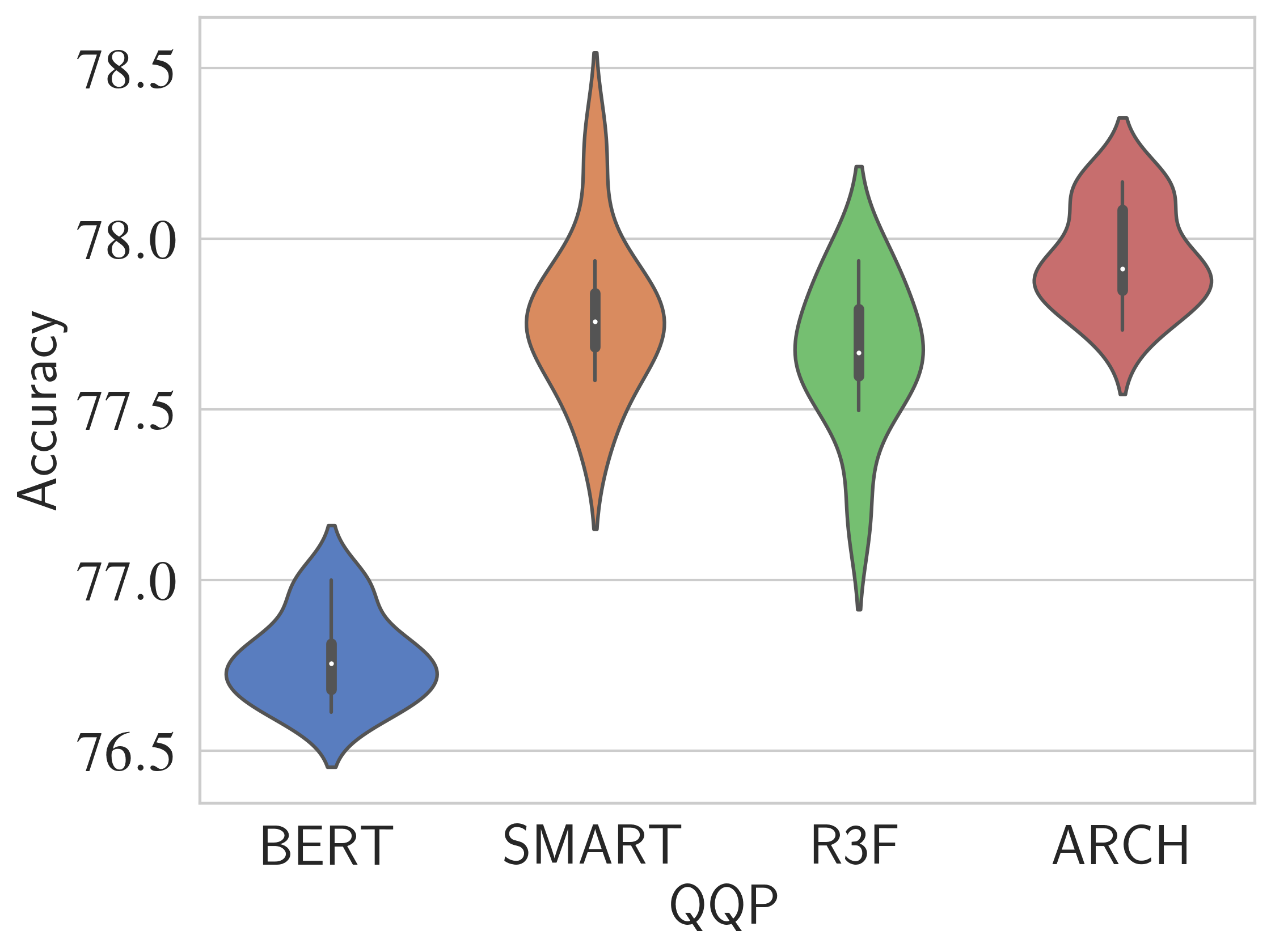}
    \vskip -0.05in
    \caption{Probing experiments. Each violin plot is based on 10 runs with different random seeds.}
    \label{fig:probing}
    % \vskip -0.1in
\end{figure}

%%%%%%%%%%%%%%%%%%%%%%%%%%%%%%
\vspace{0.05in}
\noindent $\diamond$
\textbf{Probing experiments.}
We first fine-tune a BERT\textsubscript{BASE} model on the SST-2 dataset using different methods, and then we freeze the representations and only tune a prediction head on other datasets. The probing method directly measures the quality of representations generated by different models. As illustrated in Fig.~\ref{fig:probing}, \ours consistently outperforms the baseline methods.

%%%%%%%%%%%%%%%%%%%%%%%%%%%%%%
% \vspace{0.1in}
% \noindent $\diamond$
% \textbf{Parameter study.}
% We conclude the following based on the results in Fig.~\ref{fig:analysis}:
% (1) Moving average on the perturbations helps. As indicated in the first figure, removing the moving average decreases model performance by about 0.3 BLUE.
% (2) \ours is not sensitive to the number of neighbors. This is indicated in the second figure. Notice that using a random neighbor instead of the nearest neighbor hurts model performance, and also exhibits large variance. 
% (3) \ours is robust to the proportion of cached samples, as demonstrated in the third figure.
% (4) Number of epochs between caching is important. If we cache the perturbations too frequently (i.e., 5 in the fourth figure), the model cannot adapt to the perturbations well; and if we cache the perturbations too infrequently (i.e., inf in the fourth figure), staleness of the perturbations hinders model generalization.
% We remark that all the variants of \ours in Fig.~\ref{fig:analysis} outperform the SMART baseline.

%% file: 0-conclusion.tex
\section{Conclusion}

We propose a new caching method to speedup the training of neural models with adversarial regularization. By reusing the generated perturbations, our proposed method significantly amortizes the computational cost of the backward passes at each iteration. Our thorough experiments show that the proposed method not only improves the computational efficiency, but also reduces the variance of the stochastic gradients, which leads to better model generalization.

%% file: 0-appendix.tex
%%%%%%%%%%%%%%%%%%%%%%%%%%%%%%%%%%%%%%%%%%%%%%%%%%%%%%%%%%%%
\section{Detailed Algorithm}

\begin{algorithm}[!h]
\SetAlgoLined
\caption{Adversarial Regularization with Caching and Memory Saving.}
\label{alg:full}
\KwIn{$W$: word embedding matrix from a pre-trained model; $T$: number of training epochs; $T_c$: number of epochs between caching; $\alpha$: moving average parameter.}
\tcp{Before training}
Compute $\{v_i\}_{i=1}^n$ using Eq.~\ref{eq:sentence-vector}\;
Sample a cache set $\cX \subset \{1,\cdots,n\}$\;
\For{$i \notin \cX$}{
    Find $\cK_i \subset \cX$ for $x_i$ based on cosine similarity among $\{v_i\}_{i=1}^n$\;
}
\tcp{During training}
\textbf{Initialize:} Cache $\cC=\text{dict}\{\}$\;
\For{$t=0, \cdots T-1$}{
    \For{each $x_i \in \cB$ in a batch $\cB$}{
        \uIf{$t \% T_c == 0$}{
            Find $\delta_i^t$ for each $x_i$ using projected gradient ascent\;
            \If{$i \in \cX$}{
                $\cC[x_i] \leftarrow \alpha \cC[x_i] + (1-\alpha) \delta_i^t$\;
            }
        }
        \Else{
            \uIf{$i \in \cX$}{
                $\delta_i^t=\cC[x_i]$\;
            }
            \Else{
                Compute $\delta_i^t$ using $\cK_i$, $\cC$, and Eq.~\ref{eq:construct-adv}\;
            }
        }
        One-step gradient descent on Eq.~\ref{eq:cache-loss} to update model parameters\;
    }
}
\KwOut{Trained model.}
\end{algorithm}

%%%%%%%%%%%%%%%%%%%%%%%%%%%%%%%%%%%%%%%%%%%%%%%%%%%%%%%%%%%%
\section{Training Details}

%%%%%%%%%%%%%%%%%%%%
\subsection{Machine Translation Experiments}
\label{app:nmt}

For the low-resource experiments, we use a batch size of 64k tokens. For example, when running the experiments on 4 GPUs, we set the tokens-per-GPU to be 8k, and we accumulate gradients for 2 steps. We use Adam~\cite{kingma2014adam} as the optimizer, and we set $\beta=(0.9, 0.98)$. The learning rate is set to be $1 \times 10^{-3}$ in all the experiments. We choose the model with the best validation performance to test on the test set.
Other training details are the same as \citet{ott2019fairseq}\footnote{\url{https://github.com/pytorch/fairseq/blob/master/examples/translation/README.md}}.

For the rich resource experiments, we use a batch size of 450k tokens. That is, we set tokens-per-GPU to be 7k with 8 GPUs, and we further accumulate gradients for 8 steps. We set the learning rate to be $1 \times 10^{-3}\}$. For other training setups, please refer to \citet{ott2018scaling}\footnote{\url{https://github.com/pytorch/fairseq/blob/master/examples/scaling_nmt/README.md}}.

To implement our proposed method, we sample the initial perturbation from a uniform distribution. We use sentence-level $\ell_2$ constraints on the perturbations, and we set the perturbation strength $\epsilon=0.1$. We run a modified version of projected gradient ascent for 3 steps to compute the perturbations, and the learning rate is set to be $0.1$.
Concretely, in each iteration to compute the perturbations, we apply the following update rule
\begin{align*}
    \delta \leftarrow \Pi \left( \delta + \eta \frac{\nabla_\delta \ell_v(x,\delta,\theta)}
    {\norm{\nabla_\delta \ell_v}_2} \right),
\end{align*}
where $\eta$ is the learning rate and $\Pi$ denotes the projection into the $\ell_2$ ball.
We set the number of epochs between caching to be 15, and the exponential moving average parameter $\alpha=0.01$. We cache 10\% of perturbations, and we use the nearest neighbor (i.e., 1-NN) to construct uncached perturbations.

Inference settings are presented in Table~\ref{tb:mt-parameter}.

\begin{table}[h!]
\centering
\begin{tabular}{l|cc}
\toprule
& Beam & Len-Pen \\ \midrule
En-Vi (IWSLT'15) & 10 & 1.0 \\
Vi-En (IWSLT'15) & 15 & 0.3 \\
En-De (IWSLT'14) & 10 & 1.5 \\
De-En (IWSLT'14) & 9 & 1.5 \\
En-Fr (IWSLT'16) & 10 & 0.2 \\
Fr-En (IWSLT'16) & 10 & 2.0 \\ \hline
En-De (WMT'16) & 4 & 0.6 \\
\bottomrule
\end{tabular}
\vskip -0.05in
\caption{Hyper-parameters for machine translation. Here, \textit{Beam} is the size of beam search, and \textit{Len-Pen} is the length penalty parameter during beam search.}
\label{tb:mt-parameter}
\end{table}

% \begin{table*}[t!]
% \centering
% \begin{tabular}{l|ccccc}
% \toprule
%  & batch & lr & \# epochs & dropout & BERT-dropout \\ \midrule
% RTE & 16 & $1\times 10^{-4}$ & 10 & 0.1 & 0.0 \\
% MRPC & 8 & $2\times 10^{-4}$ & 10 & 0.1 & 0.1 \\
% CoLA & 16 & $5\times 10^{-5}$ & 10 & 0.1 & 0.1 \\
% SST-2 & 32 & $8\times 10^{-5}$ & 6 & 0.1 & 0.1 \\
% STS-B & 32 & $2\times 10^{-4}$ & 10 & 0.1 & 0.0 \\
% QNLI & 32 & $1\times 10^{-4}$ & 6 & 0.1 & 0.1 \\
% QQP & 64 & $2\times 10^{-4}$ & 10 & 0.1 & 0.1 \\
% MNLI & 128 & $1\times 10^{-4}$ & 10 & 0.1 & 0.1 \\
% \bottomrule
% \end{tabular}
% \caption{Hyper-parameters for training the GLUE tasks. Here \textit{dropout} is the dropout rate of the task-specific layers (e.g., the classification heads), and \textit{BERT-dropout} is the dropout rate of the pre-trained BERT model.}
% \label{tb:hyper-glue}
% \end{table*} 

\begin{table*}[!t]
	\begin{center}
		\begin{tabular}{l|l|c|c|c|c|c}
			\toprule 
			\bf Corpus &Task& \#Train & \#Dev & \#Test   & \#Label &Metrics\\ \midrule
			\multicolumn{6}{@{\hskip1pt}r@{\hskip1pt}}{Single-Sentence Classification (GLUE)} \\ \hline
			CoLA & Acceptability&8.5k & 1k & 1k & 2 & Matthews corr\\ \hline
			SST & Sentiment&67k & 872 & 1.8k & 2 & Accuracy\\ \midrule
			\multicolumn{6}{@{\hskip1pt}r@{\hskip1pt}}{Pairwise Text Classification (GLUE)} \\ \hline
			MNLI & NLI& 393k& 20k & 20k& 3 & Accuracy\\ \hline
            RTE & NLI &2.5k & 276 & 3k & 2 & Accuracy \\ \hline
            % WNLI & NLI &634& 71& 146& 2 & Accuracy \\ \hline
			QQP & Paraphrase&364k & 40k & 391k& 2 & Accuracy/F1\\ \hline
            MRPC & Paraphrase &3.7k & 408 & 1.7k& 2&Accuracy/F1\\ \hline
			QNLI & QA/NLI& 108k &5.7k&5.7k&2& Accuracy\\ \midrule
			\multicolumn{5}{@{\hskip1pt}r@{\hskip1pt}}{Text Similarity (GLUE)} \\ \hline
			STS-B & Similarity &7k &1.5k& 1.4k &1 & Pearson/Spearman corr\\ \bottomrule
%			\multicolumn{6}{@{\hskip1pt}r@{\hskip1pt}}{Pairwise Text Classification} \bottomrule %\\ \hline
% 			SNLI & NLI& 549k &9.8k&9.8k&3& Accuracy\\ \hline
% 			SciTail & NLI& 23.5k &1.3k&2.1k&2& Accuracy\\ \hline
% 			ANLI & NLI& 163k &3.2k&3.2k&3& Accuracy\\ \hline
		\end{tabular}
	\end{center}
	\vskip -0.05in
	\caption{Summary of the GLUE benchmark.}
	\label{tb:glue}
\end{table*}

%%%%%%%%%%%%%%%%%%%%
\subsection{Natural Language Understanding Experiments}
\label{app:glue}

Statistics and descriptions of the GLUE benchmark is summarized in Table~\ref{tb:glue}.

We fine-tune a pre-trained BERT\textsubscript{BASE} model. For each task, we choose the batch size from $\{8,16,32,64,128\}$, and the learning rate from $\{5\times 10^{-5}, 8\times 10^{-5}, 1\times 10^{-4}, 2\times 10^{-4}\}$. We use a linear learning rate warm-up schedule for $10\%$ of the training iterations. We set the dropout rate of the task specific layer (i.e., the classification head) to be $0.1$, and the dropout rate of BERT is chosen from $\{0.0, 0.1\}$. We train the model for $10$ epochs. We report the best performance on each dataset individually.

To implement the adversarial regularization method, we sample the initial perturbation from a normal distribution with mean $0$ and standard deviation $10^{-5}$.
We use word-level $\ell_\infty$ constraints, and the perturbation strength is set to be $1.0$.
We run standard projected gradient ascent to compute the perturbations, where the number of steps is chosen from $\{1,2\}$, and the learning rate is chosen from $\{10^{-4}, 10^{-5}\}$.
Because of the limited number of training samples, we only cache the perturbations once for fine-tuning tasks. We refer to the \textit{MT-DNN} code-base\footnote{\url{https://github.com/namisan/mt-dnn}} for other details.